# Uzbek Cyrillic-Latin-Cyrillic Machine Transliteration

**B. Mansurov** and **A. Mansurov**
Copper City Labs
{b,a}mansurov@coppercitylabs.com

January 13, 2021

**Abstract**

In this paper, we introduce a data-driven approach to transliterating Uzbek dictionary words from the Cyrillic script into the Latin script, and vice versa. We heuristically align characters of words in the source script with sub-strings of the corresponding words in the target script and train a decision tree classifier that learns these alignments. On the test set, our Cyrillic to Latin model achieves a character level micro-averaged $F_1$ score of 0.9992, and our Latin to Cyrillic model achieves the score of 0.9959. Our contribution is a novel method of producing machine transliterated texts for the low-resource Uzbek language.

## 1 Introduction

The Uzbek language is a low-resource language, with two currently active writing systems — Cyrillic and Latin. Publicly available data for Natural Language Processing (NLP) is either in the Cyrillic script or in the Latin script, but rarely in both, if ever. The progress of NLP in the language is partly hindered by this very fact. For example, in order to build a language model, we can only utilize a subset of the available data because of the writing system of our choice. One way to solve the data scarcity issue is to transliterate available data from one writing system to the other.

Arbabi et al. 1994 describe transliteration as the process of formulating a representation of words in one language using the alphabet of another language. Alam and ul Hussain 2017 think of transliteration as converting texts written in one alphabet of a language into another alphabet of the same language. In this paper, we adopt the latter definition and tackle the issue of converting Uzbek words written in Cyrillic into words written in Latin, and vice versa. To the best of our knowledge, no such publicly available work has been done before.

### 1.1 Goals of the Paper

The goal of the paper is to present a data-driven approach to transliterating Uzbek words written in the Cyrillic script into the same words written in the Latin script and do the conversion in the opposite direction. For example, we want to build a model that can transliterate `цирк` (circus) in Cyrillic into `sirk` in Latin; and another model that is able to transliterate `sirt` (surface) in Latin into `сирт` in Cyrillic. As is evident from these examples, the task is not trivial because `s` in Latin can be transliterated as either ц or с in Cyrillic, without any apparent conversion rules. We show that, using a parallel orthography dictionary, we can correctly transliterate many words. We also discuss various ways of further improving our proposed approach.

## 1.2 Previous Work

Machine transliteration has been studied in the literature extensively. Arbabi et al. 1994 tackle the issue of transliterating Arabic names into English using a combination of neural networks and rule-based expert systems. Knight and Graehl 1997 use weighted finite-state acceptors and transducers to transliterate English words written in Japanese (katakana) back into English. Both papers employ phonetic representation of words to achieve their goals. In transliterating names from Arabic into English, Al-Onaizan and Knight 2002 present a spelling-based model using finite-state machines and achieve better results than the earlier state-of-the-art phonetic-based models.

Deselaers et al. 2009 introduce deep belief networks to transliterate Arabic names into English. Although their results were interesting, their model was not as competitive as the state-of-the-art models. Alam and ul Hussain 2017 create a sequence-to-sequence model with three layers of long short-term memory (LSTM) based encoder and decoder to transliterate Roman-Urdu words to Urdu and achieve Bilingual Evaluation Understudy (BLEU) score of 48.6 on the test set. Le and Sadat 2018 use recurrent neural networks (RNNs) to transliterate words between French and Vietnamese using their phonetic representation.

Najafi et al. 2018 do a comparative study of different transliteration methods on NEWS Shared Task on Machine Transliteration and find that "… on average, the neural models perform better than other systems, and that a combination of neural and non-neural models further improves the results".

Most works in the literature deal with a somewhat difficult task of transliterating words from one language into another. In this paper, our task is relatively simple as we only deal with a single language — Uzbek. Our solution is also simple and unique (as far as we know) with promising results.

## 2 Methods

### 2.1 Background knowledge

The Cyrillic alphabet of the Uzbek language consists of the following 35 letters and their lowercase variants: А, Б, В, Г, Д, Е, Ё, Ж, З, И, Й, К, Л, М, Н, О, П, Р, С, Т, У, Ф, Х, Ц, Ч, Ш, Ъ, Ь, Э, Ю, Я, Ў, Қ, Ғ, and Ҳ. The Latin alphabet consists of the following 30 letters and their lowercase variants: A, B, D, E, F, G, H, I, J, K, L, M, N, O, P, Q, R, S, T, U, V, X, Y, Z, O', G', Sh, Ch, Ng, and '. To give you a glimpse of how these two alphabets map to each other, consider the following case. The letter A in Cyrillic can only appear as A in Latin, while the letter A in Latin can appear as either A or Я in Cyrillic.

Although conversion rules from Cyrillic into Latin exist, information is irrecoverably lost during conversion. For example, октябрь (October) in Cyrillic is transliterated as oktabr in Latin. Notice how the Cyrillic letter ь has no equivalent in Latin — it just does not exist in the converted text. If we follow an imaginary letter-by-letter Latin to Cyrillic conversion rule, we will end up with an incorrect transliteration of oktabr in Latin into октабр in Cyrillic, which has one wrong and one missing character from the correct transliteration: октябрь.

The orthography rules based on the Cyrillic script were approved on April 4, 1956, while the rules based on the Latin script were approved on August 24, 1995 [Tog'ayev et al. 1999]. A new effort to improve the existing Latin alphabet started in 2018[1]. Here are some of the differences between the two orthography rules:

- In Cyrillic many loanwords are written according to their spelling in the foreign language, but in Latin they are written according to their pronunciation: октябрь → oktabr (October), ноябрь → noyabr (November), бюджет → budjet (budget).
- When the Cyrillic letter ц appears as the first or last letter of a word, it is written as s in Latin: цемент → sement (cement) and шприц → shpris (syringe). Inside a word, when ц appears after a vowel, it's written as ts, but when it appears after a consonant, it's written as s: доцент → dotsent (Associate Professor), лекция → leksiya (lecture).
- When the suffix га is added to words ending in the sound ғ, in Cyrillic both г and ғ change to қ, while this change does not happen in Latin: боғ+га=боққа → bog'+ga=bog'ga (to the garden).
- Particles appearing as a conjunction between words are written with a hyphen in Latin: фикру ёд → fikr-u yod (thought and memory).
- In Latin, a hyphen is written after numbers indicating dates: 2020 йил, 20 ноябрь → 2020-yil, 20-noyabr (November 20, 2020).

In the next section, we describe our approach to learning these rules from data. When talking about Cyrillic to Latin transliteration, we refer to Cyrillic words as source words, and to Latin words as target words. Similarly, when talking about Latin to Cyrillic transliteration, we refer to Latin words as source words, and to Cyrillic words as target words.

## 2.2 Approach

We solve the problem of transliteration by splitting the source word into individual characters and aligning these characters with zero or more characters of the target word. For a given source character, we create a vector that consists of its surrounding characters. Then we train a decision tree classifier using these vectors as features and the source characters' corresponding target characters as classes. We choose the best model and its hyperparameters based on the micro-averaged $F_1$ score on the validation set. Let us discuss our approach in detail below.

Transliterating a source word can be viewed as transliterating individual letters of the word so that the resulting characters put together forms the target word. We hypothesize that a letter along with its surrounding letters in a word carries enough information to be transliterated correctly. In order to test our hypothesis, first, we align each letter of the source word with zero or more letters of the target word. For example, Table 1 shows one such alignment where the source word is қўзичоқ (lambkin) in Cyrillic. Notice how the source word is split into individual characters that align with both single (indices 0, 2, 3, 5, and 6) and double characters (indices 1 and 4) of the target word.

| Index | 0 | 1 | 2 | 3 | 4 | 5 | 6 |
|---|---|---|---|---|---|---|---|
| Cyrillic source word | қ | ў | з | и | ч | о | қ |
| Latin target word | q | o' | z | i | ch | o | q |

Table 1: An alignment of қўзичоқ (lambkin) in Cyrillic with its Latin equivalent.

---

[1]http://uza.uz/oz/society/lotin-yezuviga-asoslangan-zbek-alifbosi-a-ida-ishchi-guru-ni-06-11-2018

Table 2 shows a similar alignment, but in this case the word in Cyrillic is the target word. Here again, the source word is split into individual characters, while the target word is split into zero (indices 2 and 5) or more (indices 0, 1, 3, 4, 6, 7, and 8) characters.

| Index | 0 | 1 | 2 | 3 | 4 | 5 | 6 | 7 | 8 |
|---|---|---|---|---|---|---|---|---|---|
| Latin source word | q | o | ' | z | i | c | h | o | q |
| Cyrillic target word | қ | ў | | з | и | | ч | о | қ |

Table 2: An alignment of `qo'zichoq` (lambkin) in Latin with its Cyrillic equivalent.

A natural question arises: why do we have to split the source word into individual letters? After all, the Cyrillic letter ч is written as `ch` in Latin and not as `h`. One of the reasons is that some Latin letters consist of a single character (e.g., `A`), while others consist of two characters (e.g., `YA`). Splitting the source word at the individual character level makes training a model and using it to transliterate words easier.

Consider the word `rayon` (district) in Latin. Its alignment is shown in Table 3. Each Latin letter has a corresponding Cyrillic variant. Now consider a multi-character alignment of the word `quyosh` (the Sun) in Latin shown in Table 4. In this case the Latin characters `y` and `o` combine to make the Cyrillic letter ё. Rather than trying to identify whether two Latin characters align with one or two Cyrillic characters, we let our algorithm learn these mappings from data. The only thing we must do is to construct Cyrillic to Latin and Latin to Cyrillic mappings of characters.

| Index | 0 | 1 | 2 | 3 | 4 |
|---|---|---|---|---|---|
| Latin source word | r | a | y | o | n |
| Cyrillic target word | р | а | й | о | н |

Table 3: An alignment of `rayon` (district) in Latin with its Cyrillic equivalent.

| Index | 0 | 1 | 2 | 3 |
|---|---|---|---|---|
| Latin source word | q | u | yo | sh |
| Cyrillic target word | қ | у | ё | ш |

Table 4: An alignment of `quyosh` (the Sun) in Latin with its Cyrillic equivalent.

Table 5 shows Cyrillic to Latin mappings and Table 6 shows Latin to Cyrillic mappings. We learn these mappings heuristically. From previous experience working with both scripts, we know that the Cyrillic letter Б always maps to the Latin letter B. In fact, we know that each of the 25 (out of 36) characters in Table 5 map to only one letter. We identify the remaining mappings by going over the words in the dictionary and trying to look up a mapping for each character from Table 5. If no such mapping exists, we manually examine the word in question and add the corresponding mapping to the table. That way we fill Table 5 with the remaining mappings. To fill Table 6, we repeat the above process with words in Latin as source words and words in Cyrillic as target words.

Once we have filled both tables, we loop over our data and create alignments like the ones seen in Table 1 and Table 2. In essence, our problem is now reduced to a multi-class classification problem: we have a finite number of source characters that map to a finite number of target strings and we need to figure out which source characters map to which target strings. Many algorithms such as a Naive Bayes or a decision tree classifier can be used to tackle this problem.

To train a classifier we start with an alignment and consider the source characters as features and their corresponding target mappings as classes. We gather all such features and classes and feed them into our classifier. For example, we create seven data points (one for each letter) using the Cyrillic characters of the word қўзичоқ (lambkin).

| - → -   | И → I,YI,U | С → S    | Ь → ∅    |
|---------|------------|----------|----------|
| А → A   | Й → Y      | Т → T    | Э → E    |
| Б → B   | К → K      | У → U,-U | Ю → U,YU |
| В → V   | Л → L      | Ф → F    | Я → A,YA |
| Г → G   | М → M      | Х → X    | Ё → YO,O |
| Д → D   | Н → N      | Ц → S,TS | Ў → O',∅ |
| Е → E,YE| О → O,YO   | Ч → CH,∅ | Ғ → G'   |
| Ж → J   | П → P      | Ш → SH   | Қ → Q    |
| З → Z   | Р → R      | Ъ → ',∅  | Ҳ → H    |

Table 5: Cyrillic to Latin character mappings. ∅ denotes an empty string.

| - → -    | G → Г,Ғ   | N → НЬ,Н       | U → И,У,Ю   |
|----------|-----------|----------------|-------------|
| A → А,Я  | H → Х,Ш,Ч | O → О,Ё,Ўъ,Ў   | V → ВЬ,В    |
| B → БЪ,Б | I → ЧИ,И  | P → ПЬ,П       | X → Х       |
| C → ∅    | J → Ж     | Q → Қ          | Y → Й,∅     |
| D → ДЬ,Д | K → К     | R → РЬ,Р       | Z → ЗЪ,ЗЬ,З |
| E → Е,Э  | L → ЛЬ,Л  | S → СЬ,С,Ц,    | ' → ∅       |
| F → ФЬ,Ф | M → МЬ,М  | T → ТЬ,Т,∅     | ' → Ъ,∅     |

Table 6: Latin to Cyrillic character mappings. ∅ denotes an empty string.

Since not all characters map one-to-one to classes (as seen in Table 5 and Table 6), we also consider features consisting of the original character and its surrounding characters. For a given character in the source word, we create a vector of features consisting of X number of characters before it and Y number of characters after it. X and Y are the hyperparameters of our model. If a word is shorter than the desired number of preceding or subsequent characters, we pad the word with the ∅ (empty set) character. To give you an example, let the number of preceding characters be two, and the number of subsequent characters be one. The feature vectors and classes of the letters of қўзичоқ (lambkin) are shown in Table 7.

| Feature 1 | Feature 2 | Feature 3 | Feature 4 | Class |
|-----------|-----------|-----------|-----------|-------|
| ∅         | ∅         | қ         | ў         | q     |
| ∅         | қ         | ў         | з         | o'    |
| қ         | ў         | з         | и         | z     |
| ў         | з         | и         | ч         | i     |
| з         | и         | ч         | о         | ch    |
| и         | ч         | о         | қ         | o     |
| ч         | о         | қ         | ∅         | q     |

Table 7: Feature vectors and classes of the letters of қўзичоқ (lambkin) with two preceding and one subsequent character.

## 2.3 Data

The words in Cyrillic and their Latin variants used in our experiments are taken from an Uzbek Cyrillic to Latin spelling dictionary [Tog'ayev et al. 1999]. The dictionary, first and foremost, includes widely used words in the contemporary Uzbek literary language. Its authors tried not to include words that are straightforward to spell or words whose derivational affixes are simple. However, words that are prone to spelling errors are included in the dictionary.

The dictionary consists of 13,855 entries, and it does not include abbreviations, acronyms, proper nouns, or words with morphological inflection (except for a few cases). After removing multi-word phrases and entries with punctuation marks, we are left with 12,418 words. These words are shuffled randomly and split into three sets: 9,499 words for training, 1,677 words for validation, and 1,242 words for testing.

## 2.4 Experiments

We experimented with a Naive Bayes and a decision tree classifier. Our model based on the decision tree classifier had the highest scores. We think that the naivety assumption of the Naive Bayes classifier prevented the model from learning important letter sequences. On the other hand, the decision tree classifier is learning these sequences of letters due to our features being sequential too. We are not worried about overfitting because changes in language happen over time and the number of words and affixes are limited in the short term.

We used `scikit-learn`'s[2] implementation of the decision tree classifier with the default parameters[3] to carry out our experiments. In order to find the best hyperparameters, we created feature vectors consisting of a combination of zero to ten preceding and subsequent characters. For each combination of hyperparameters we converted the train, validation, and test datasets into feature vectors and classes. We also removed duplicate data points (within datasets) from all datasets. We then trained and picked our best models based on the character level micro-averaged $F_1$ scores on the validation set.

## 3 Results

The hyperparameters of our best models are shown in table 8. To briefly explain our Cyrillic to Latin model, it works best when, in addition to the letter that is being transliterated, it has access to two previous and three subsequent characters of the word. For example, in order to transliterate the Cyrillic я in октябрь (October) into the Latin a (and not ya), the model makes the best decision when it knows the two preceding letters кт and the three subsequent letters брь.

|  | # of preceding characters | # of subsequent characters |
|---|---|---|
| Cyrillic to Latin model | 2 | 3 |
| Latin to Cyrillic model | 4 | 3 |

Table 8: Best model hyperparameters.

---

[2] https://scikit-learn.org/
[3] criterion="gini", splitter="best", max_depth=None, min_samples_split=2, min_samples_leaf=1, min_weight_fraction_leaf=0.0, max_features=None, random_state=None, max_leaf_nodes=None, min_impurity_decrease=0.0, min_impurity_split=None, class_weight=None, ccp_alpha=0.0

The character level scores of our best models on the test set are shown in Table 9. The Latin to Cyrillic model is achieving lower scores because the existing conversion rules are designed to convert Cyrillic to Latin, and not Latin to Cyrillic (such rules do not exist for Latin to Cyrillic). As such, we can better capture the existing rules for converting Cyrillic to Latin.

|  | Precision | Recall | Micro-averaged $F_1$ score |
| --- | --- | --- | --- |
| Cyrillic to Latin model | 0.9992 | 0.9992 | 0.9992 |
| Latin to Cyrillic model | 0.9959 | 0.9959 | 0.9959 |

Table 9: Character level scores of our best models on the test set.

Tables 10 and 11 show the words that our classifiers made a mistake on. Both models are making these errors because of the lack of similar data in training. For example, there is only one word that contains the letters фью in our dictionary, and that word only appears in the test dataset. That is why our model was not able to correctly transliterate the word фьючерс (futures).

| Model Input | Model Output | Correct Transliteration | English Translation |
| --- | --- | --- | --- |
| фьючерс | fuchers | fyuchers | futures |
| итялоқ | italoq | ityaloq | dog bowl |
| култилламоқ | qultillamoq | qultullamoq | swallow |
| мешчан | meshan | meshchan | petty bourgeois |
| хусусийлаштириш | xususuylashtirish | xususiylashtirish | privatization |
| эшакеми | eshakemi | eshakyemi | a type of skin disease |

Table 10: Errors made by our best Cyrillic to Latin model. Six words out of 1,242 were transliterated incorrectly.

## 4 Discussion

In this paper, we shared our attempt at building a transliterator of Uzbek dictionary words from Cyrillic to Latin, and vice versa. Our models achieve remarkably high character level micro-averaged $F_1$ scores on the test set. However, our training data consists of only lowercase words, and it does not include running texts, abbreviations, acronyms, proper nouns, words with morphological inflection (except for a few cases), words consisting of all capitals (as in headlines), other types of words not mentioned here, or punctuation marks. It would be interesting to see how our approach handles these additional types of data.

To build a production quality transliterator, we also need to take user input inconsistencies into account. For example, the Latin letter o' is often written in one of the following forms o', o`, ó, o', and o′, only one of which is correct. Cleaning and fixing user input before feeding it into the transliterator may be necessary. Even then, our transliterator will not produce perfect results. In such cases analyzing and correcting the output of the transliterator using a high-quality language model may help.

## 5 Acknowledgements

We would like to thank N. Mansurov for manually verifying and fixing our training, validation, and test datasets.

| Model Input | Model Output | Correct Transliteration | English Translation |
| --- | --- | --- | --- |
| valuta | валута | валюта | currency |
| kvadrat | квядрат | квадрат | square |
| piltakach | пильтакач | пилтакач | tailor needle with a small handle |
| marsiya | марция | марсия | memorial poem |
| xususiyat | хусуцият | хусусият | feature |
| kastrulka | каструлка | кастрюлька | pan |
| apellatsiya | апеллация | апелляция | appeal |
| spektakl | спектакл | спектакль | show |
| rukzak | рукзак | рюкзак | backpack |
| gʻudullamoq | ғудилламоқ | ғудулламоқ | mumble |
| yakor | якор | якорь | anchor |
| federal | федераль | федерал | federal |
| etikdoʻz | этикдӱзь | этикдӱз | cobbler |
| qitmir | қитмирь | қитмир | schemer |
| ziravor | зирявор | зиравор | seasoning |
| medal | медал | медаль | medal |
| konferansye | конферансе | конферансье | master of ceremonies |
| ekskursiya | экскурция | экскурсия | excursion |
| monastir | монастир | монастирь | monastery |
| feldmarshal | фелдмаршал | фельдмаршал | field-marshal |
| golf | голф | гольф | golf |
| plyonka | плонка | плёнка | film |
| aksiz | аксиз | акциз | excise |
| yengil | енгиль | енгил | light |
| kilkillamoq | килькилламоқ | килкилламоқ | shiver |
| basseyn | басцейн | бассейн | pool |
| sellofan | целлофан | селлофан | cellophane |
| tavsiya | тавция | тавсия | recommendation |
| karusel | карусел | карусель | carousel |
| diagonal | диагонал | диагональ | diagonal |
| vilt | вильт | вилт | wilt |
| porshen | поршен | поршень | piston |
| sanksiya | санькция | санкция | sanction |
| valeryanka | валерянка | валерьянка | valerian drops |

Table 11: Errors made by our best Latin to Cyrillic model. Thirty-four words out of 1,242 were transliterated incorrectly.